\def\year{2022}\relax
\documentclass[letterpaper]{article} % DO NOT CHANGE THIS
\usepackage{aaai22}  % DO NOT CHANGE THIS
\usepackage{times}  % DO NOT CHANGE THIS
\usepackage{helvet}  % DO NOT CHANGE THIS
\usepackage{courier}  % DO NOT CHANGE THIS
\usepackage[hyphens]{url}  % DO NOT CHANGE THIS
\usepackage{graphicx} % DO NOT CHANGE THIS
\usepackage{natbib}  % DO NOT CHANGE THIS AND DO NOT ADD ANY OPTIONS TO IT
\usepackage{caption} % DO NOT CHANGE THIS AND DO NOT ADD ANY OPTIONS TO IT
\DeclareCaptionStyle{ruled}{labelfont=normalfont,labelsep=colon,strut=off} % DO NOT CHANGE THIS
\usepackage{algorithm}
\usepackage{algorithmic}

\usepackage{newfloat}
\usepackage{listings}
\floatstyle{ruled}
\newfloat{listing}{tb}{lst}{}
\floatname{listing}{Listing}
\usepackage[sfdefault,lf]{carlito}
\usepackage[T1]{fontenc}

\usepackage[utf8]{inputenc}
\usepackage{xcolor}
\usepackage{booktabs}
\usepackage{amsthm}
\usepackage{listings}
\usepackage{inconsolata}
\usepackage{tikz}
\usepackage{pgfplots}
\usepackage{amsmath}
\usepackage{microtype}
\usepackage{algorithm}

\usepackage[normalem]{ulem} % for \sout
\usepackage{subcaption}
\usepackage[misc]{ifsym} % for \Letter
\usepackage{tikz}
\usetikzlibrary{calc}
\usetikzlibrary{arrows}
\usetikzlibrary{shapes}
\usetikzlibrary{decorations}
\usetikzlibrary{arrows.meta}
\definecolor{pixel 0}{HTML}{FFFFFF}
\definecolor{pixel 1}{HTML}{FF0000} % red

\newcommand{\tw}[1]{\texttt{#1}}
\newcommand{\name}{\textsc{Popper}}
\newcommand{\popper}{\textsc{Popper}}
\newcommand{\parname}{\textsc{ParSearch}}

\newcommand{\dac}{D\&C}
\newcommand{\dacc}{D\&C$_{com}$}
\newcommand{\port}{Portfolio}
\newcommand{\portc}{Portfolio$_{com}$}

\usepackage{pgfkeys}
    \newenvironment{customlegend}[1][]{%
        \begingroup
        \csname pgfplots@init@cleared@structures\endcsname
        \pgfplotsset{#1}%
    }{%
        \csname pgfplots@createlegend\endcsname
        \endgroup
    }%
    \def\addlegendimage{\csname pgfplots@addlegendimage\endcsname}

\theoremstyle{definition}
\newtheorem{definition}{Definition}

\newenvironment{code}
{
\ttfamily
\begin{center}
\begin{tabular}{l}
}
{
\end{tabular}
\end{center}
\par
}

\lstnewenvironment{myalgorithm}[1][] %defines the algorithm listing environment
{
    \lstset{ %this is the stype
        showspaces=false,               % show spaces adding particular underscores
        showstringspaces=false,         % underline spaces within strings
        mathescape=true,
        numbers=left,
        escapeinside={*}{*},
        columns=flexible,
        numbersep=4pt,        % default 10pt
        keywordstyle=\bfseries,
        keywords={,and, return, not, def, in, if, else, for, foreach, while, }
        numbers=left,
        xleftmargin=.01\textwidth,
        #1 % this is to add specific settings to an usage of this environment (for instnce, the caption and referable label)
    }
}
{}

\title{Parallel Constraint-Driven Inductive Logic Programming}
\author{
    Andrew Cropper,\textsuperscript{\rm 1}
    Oghenejokpeme Orhobor, \textsuperscript{\rm 2}
    Cristian Dinu, \textsuperscript{\rm 1}
    Rolf Morel \textsuperscript{\rm 1} \\
}
\affiliations{
    \textsuperscript{\rm 1} University of Oxford \\
    \textsuperscript{\rm 2} University of Cambridge \\
    andrew.cropper@cs.ox.ac.uk, oo288@cam.ac.uk, cristian.dinu@hertford.ox.ac.uk, rolf.morel@cs.ox.ac.uk
}

\begin{document}
% \nocopyright
\maketitle

\begin{abstract}
Multi-core machines are ubiquitous.
However, most inductive logic programming (ILP) approaches use only a single core, which severely limits their scalability.
To address this limitation, we introduce parallel techniques based on \emph{constraint-driven} ILP where the goal is to accumulate constraints to restrict the hypothesis space.
Our experiments on two domains (program synthesis and inductive general game playing) show that (i) parallelisation can substantially reduce learning times, and (ii) worker communication (i.e. sharing constraints) is important for good performance.
\end{abstract}

\section{Introduction}
Inductive logic programming (ILP) \cite{mugg:ilp} is a form of machine learning.
Given positive and negative examples of a predicate and background knowledge (BK), the ILP problem is to find a set of logical rules (a \emph{hypothesis}), which, with the BK, entails the positive examples and none of the negative examples.
Key features of ILP include its ability to (i) learn from small amounts of data, (ii) support relational data, and (iii) induce human-readable hypotheses \cite{ilp30}.
% \rolf{Is this the most appropriate reference here? The main thing I am thinking of here is the large number of self cites that reviewers will also catch.}

Although powerful, ILP approaches often struggle with scalability: efficiently searching a large hypothesis space (the set of all hypotheses) for a \emph{solution} (a hypothesis that correctly generalises the examples).
Parallelisation is one approach to improving scalability.
However, although multi-core machines are ubiquitous, most ILP approaches are single-core learners \cite{progol,aleph,metagol,ilasp}.
 % and Aleph \cite{aleph}, employ a sequential set covering approach, where they learn individual rules to generalise examples.

To overcome this limitation, we introduce parallel techniques based on \emph{constraint-driven} ILP where the goal is to accumulate constraints to restrict the hypothesis space.
In particular, we build on the \emph{learning from failures} (LFF) \cite{popper,poppi} approach, which supports predicate invention and learning optimal and recursive programs.
A LFF learner works by repeatedly generating and testing hypotheses on training examples.
If a hypothesis fails to correctly generalise the examples, the learner deduces constraints to \emph{explain} the failure, which it then uses to rule out other hypotheses and thus restrict the hypothesis space.
The process repeats until a solution is found.

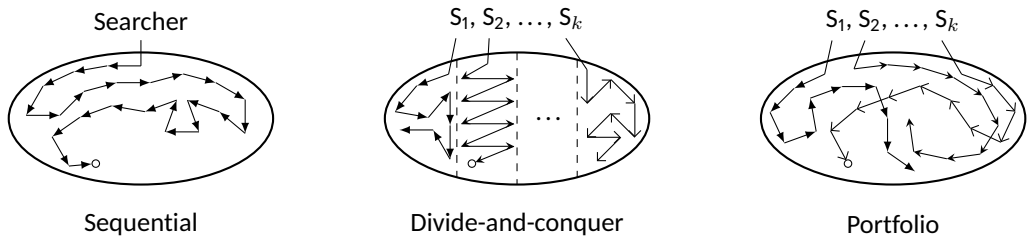
\begin{figure*}[!t]
\centering
\begin{tikzpicture}
\usetikzlibrary{arrows}
\usetikzlibrary{shapes}
\usetikzlibrary{calc}
\usetikzlibrary{decorations}
\usetikzlibrary{arrows.meta}

\pgfdeclaredecoration{multiple arrows}{draw}{%
  \state{draw}[width=\pgfdecoratedinputsegmentlength]{%
    \draw [multiple arrows path/.try] (0,0) -- (\pgfdecoratedinputsegmentlength,0);
  }
}
\tikzset{multiple arrows/.style={multiple arrows path/.style={#1},decoration=multiple arrows, decorate}}

\tikzstyle{space}=[thick,ellipse,draw=black,minimum width=100pt,minimum height=50pt,align=center]
%\tikzstyle{circ}=[circle,scale=0.1pt,draw=red]
\tikzstyle{circ}=[scale=0.1pt]
\tikzstyle{goal}=[circle,scale=0.3pt,draw=black,fill=none]

\draw 
(0,0) node[space] (space1) {}
(5,0) node[space] (space2) {}
(10,0) node[space] (space3) {};
\draw
($(space1)-(0,1.4)$) node {Sequential}
($(space2)-(0,1.4)$) node {Divide-and-conquer}
($(space3)-(0,1.4)$) node {Portfolio};

\draw[dashed] ($(space2)+(-0.8,0.78)$) -- ++(0,-1.56)
($(space2)+(0,0.88)$) -- ++(0,-1.76)
($(space2)+(0.8,0.78)$) -- ++(0,-1.56);
\draw (space2) ++(0.44,0) node {\ldots};

\draw ($(space1)-(0.6,0.6)$) node[style=goal] (G1) {}
($(space2)-(0.6,0.6)$) node[style=goal] (G2) {}
($(space3)-(0.6,0.6)$) node[style=goal] (G3) {};

\draw ($(space1)+(0,1.3)$) node (S1S1_) {Searcher}
(S1S1_.south) node[style=circ] (S1S1) {};

\draw ($(space2)+(0,1.33)$) node (S2S) {S\textsubscript{1}, S\textsubscript{2}, \ldots, S$_k$};
\draw ($(S2S)-(0.81,0.25)$) node[style=circ] (S2S1) {}
($(S2S)-(0.37,0.25)$) node[style=circ] (S2S2) {}
($(S2S)+(0.64,-0.25)$) node[style=circ] (S2SK) {};

\draw ($(space3)+(0,1.3)$) node (S3S) {S\textsubscript{1}, S\textsubscript{2}, \ldots, S$_k$};
\draw ($(S3S)-(0.81,0.25)$) node[style=circ] (S3S1) {}
($(S3S)-(0.37,0.25)$) node[style=circ] (S3S2) {}
($(S3S)+(0.64,-0.25)$) node[style=circ] (S3SK) {};

\draw[] (S1S1) edge ($(space1)+(0,0.7)$);
\path [multiple arrows={-latex}] ($(space1)+(0,0.7)$)
-- ++(180:0.45)
-- ++(190:0.45)
-- ++(205:0.45)
-- ++(240:0.45)
-- ++(0:0.45)
-- ++(45:0.45)
-- ++(15:0.45)
-- ++(0:0.45)
-- ++(15:0.45)
-- ++(-15:0.45)
-- ++(-30:0.45)
-- ++(-90:0.45)
-- ++(140:0.45)
-- ++(160:0.45)
-- ++(-70:0.45)
-- ++(180:0.45)
-- ++(70:0.45)
-- ++(200:0.45)
-- ++(170:0.45)
-- ++(200:0.45)
-- ++(215:0.45)
-- ++(-60:0.45)
-- (G1)
;

\draw[] (S2S1) edge ($(space2)+(-0.91,0.59)$);
\def\angle{0}
\path [multiple arrows={-latex}] ($(space2)+(-0.91,0.59)$)
\foreach \stepangle[evaluate=\stepangle as \angle using \angle+\stepangle, remember=\angle] in 
{200,30,120,60,-140,0,210,60}
  { -- ++(\angle:0.45) }
;
\draw[] (S2S2) edge ($(space2)+(-0.39,0.73)$);
\path [multiple arrows={-stealth}] ($(space2)+(-0.39,0.73)$)
-- ++(-0.35,-0.2)
-- ++(0.7,0)
-- ++(-0.7,-0.3)
-- ++(0.7,0)
-- ++(-0.7,-0.3)
-- ++(0.7,0)
-- ++(-0.7,-0.3)
-- ++(0.7,0)
-- (G2)
;
\draw[] (S2SK) edge ($(space2)+(0.93,0.65)$);
\path [multiple arrows={-Straight Barb}] ($(space2)+(0.93,0.65)$)
\foreach \stepangle[evaluate=\stepangle as \angle using \angle+\stepangle, remember=\angle] in 
{-90,135,-90,-45,-135,90,135,-135}
  { -- ++(\angle:0.45) }
;

\draw[] (S3S1) edge ($(space3)+(-0.91,0.59)$);
\path [multiple arrows={-latex}] ($(space3)+(-0.91,0.59)$)
\foreach \stepangle[evaluate=\stepangle as \angle using \angle+\stepangle, remember=\angle] in 
{-155,20,70,85,80,-80,-20,-70,-20,55}
  { -- ++(\angle:0.45) }
;
\draw[] (S3S2) edge ($(space3)+(-0.52,0.66)$);
\path [multiple arrows={-stealth}] ($(space3)+(-0.52,0.66)$)
\foreach \stepangle[evaluate=\stepangle as \angle using \angle+\stepangle, remember=\angle] in 
{7,-12,-12,-15,-30,-60,-50,-20,-70}
  { -- ++(\angle:0.45) }
;

\draw[] (S3SK) edge ($(space3)+(0.91,0.59)$);
%\path [multiple arrows={-stealth}] ($(space3)+(0.91,0.59)$)
%\foreach \stepangle[evaluate=\stepangle as \angle using \angle+\stepangle, remember=\angle] in 
%{-20,-30,-65,-30,-45,20,-100,20,90,20}
%  { -- ++(\angle:0.45) }
%-- (G3)
%;
\path [multiple arrows={-Straight Barb}] ($(space3)+(0.91,0.59)$)
\foreach \stepangle[evaluate=\stepangle as \angle using \angle+\stepangle, remember=\angle] in 
{-20,-30,-65,-90,-30,30,45,0,20}
  { -- ++(\angle:0.45) }
-- (G3)
;

\end{tikzpicture}
\caption{One sequential and two parallel search strategies for finding a solution in the same hypothesis space.}
\label{fig:strategies}
\end{figure*}

We introduce two general constraint-driven parallel ILP approaches based on parallel \emph{conflict-driven clause learning} (CDCL) SAT techniques \cite{parsat}, namely \emph{portfolio} and \emph{divide-and-conquer} approaches.
In our portfolio approach, parallel LFF learners compete by searching the same hypothesis space using different heuristics.
In our divide-and-conquer approach, we divide the hypothesis space into disjoint subspaces which we assign to parallel LFF learners who each search them.
Figure \ref{fig:strategies} illustrates these strategies.
We also allow learners to exchange learned constraints, similar to how parallel SAT techniques share clauses.

Overall, our contributions are:
\begin{itemize}
    \item We introduce parallel ILP approaches inspired by \emph{portfolio} and \emph{divide-and-conquer} approaches used by CDCL SAT solvers.
    \item We implement the techniques to parallelise the LFF implementation \name{}.
    \item We experimentally show on two domains (program synthesis and inductive general game playing) that (i) our parallel methods can lead to linear speedups with up to four processors in general, (ii) our parallel methods can lead to super-linear speedup in some cases, and (iii) that communication (i.e. sharing constraints) is important for good performance.
\end{itemize}

\section{Related Work}
\label{sec:related}

\subsection{Sequential ILP}
Many ILP systems, such as Progol \cite{progol} and Aleph \cite{aleph}, often struggle with large numbers of examples because of their sequential set covering approach.
A notable exception is QuickFOIL \cite{quickfoil} which builds on FOIL \cite{foil} by introducing (i) a new scoring function for clauses, and (ii) a highly efficient relational database implementation.
The authors show that their approach can scale to datasets with millions of background facts and hundreds of thousands of examples.
However, because it builds on FOIL, QuickFOIL inherits its limitations, including (i) difficulty learning recursive programs, (ii) no support for predicate invention, and (iii) no guarantees about the optimality of solutions.
These limitations apply to almost all classical ILP systems \cite{ilp30}.

Modern meta-level ILP approaches \cite{aspal,metagol,ilasp,hexmil,apperception,popper} can often learn recursive and optimal programs and perform predicate invention.
Although there is no standard definition for \emph{meta-level ILP} most approaches encode the ILP problem as a meta-level logic program, i.e.~a program that reasons about programs.
These approaches delegate the search for a hypothesis to an off-the-shelf solver, such as an answer set programming (ASP) solver \cite{clingo}, after which the meta-level solution is translated back to a standard solution for the ILP task.
However, modern meta-level ILP approaches often struggle in terms of scalability.
For instance, ILASP \cite{ilasp} struggles to learn rules with lots of literals because it precomputes every possible rule that may appear in a program, which is often infeasible.
The same issue prevents HEXMIL \cite{hexmil} from scaling to large problems.
As far as we are aware, there is no parallel meta-level ILP system.

\subsection{Parallel ILP}
\citet{DBLP:journals/ml/FonsecaSSC09} survey parallel ILP techniques.
They divide approaches into three categories: \emph{search}, \emph{data}, and \emph{evaluation}.
Search approaches parallelise the search of the hypothesis space.
Our approaches are in this category.
Data approaches divide the training examples amongst the workers which learn solutions for them in parallel.
Evaluation approaches evaluate candidates rules in parallel.

We discuss three notable parallel approaches.
\citet{parallelclaudien} parallelise the search for a hypothesis in the ILP system Claudien.
Claudien works by maintaining a priority queue of potential clauses to add to a hypothesis.
If a clause is too general, it is removed from the queue and its specialisations are added.
In the parallel approach, parallel workers process clauses in the queue.
The authors experimentally show that the parallelisation speed-up is roughly proportional to the number of workers.

\citet{wang2000parallel} parallelise Progol by allocating a subset of the positive examples (and all of the negative examples) to each worker.
Each worker applies the standard Progol sequential algorithm to find a good clause for its subset of the positive examples, which it then communicates to the other workers, who may then incorporate the clause into their local theory.
The authors show linear speedups with four and six processors.

\citet{DBLP:journals/ml/SrinivasanFJ12} use the MapReduce paradigm to parallelise the scoring step of Aleph, where a clause is generated and its score is calculated as a function of the examples.
Their results are generally positive, especially when the number of examples is large.
However, although the authors rightly claim that their approach is not specific to Aleph, it is specific to the classic set covering approach.

These three approaches are all limited because they inherit the same limitations as their sequential counterparts.
% , namely difficulty learning recursive and optimal programs and an inability to perform predicate invention.
By contrast, a key contribution of this paper is the introduction of parallel techniques that can learn optimal and recursive programs and perform predicate invention.
% \rolf{As predicate invention is mentioned multiple times, might want to add a short sentence somewhere explaining what it means.}

More recently, \citet{DBLP:conf/ilp/KatzourisAP17} introduce p-OLED, a parallel version of OLED \cite{oled},  which learns event definitions in the form of even calculus theories.
In their parallel approach, a clause is evaluated in parallel on sub-streams of the input stream and its independent scores are combined.
Their evaluations show that their approach can reduce training times and, in some cases, is capable of super-linear speed-ups.
In contrast to p-OLED, our approaches learn general definite programs, including programs with recursion and predicate invention.

\subsection{Parallel SAT}
\label{sec:parsat}

Our parallel ILP approaches are based on parallel conflict-driven-clause-learning SAT techniques \cite{parsat}, of which there are two main approaches.
\emph{Divide-and-conquer} approaches divide the search space into sub-spaces which are allocated to sequential workers.
Workers co-operate through the exchange of learnt conflicts.
Portfolio approaches \cite{manysat} allow multiple sequential workers to compete on the same search space by employing different search heuristics, such as different restart policies, branching heuristics, random seeds, etc.
Workers also co-operate through the exchange of learnt conflicts.
This exchange can be done through message passing \cite{DBLP:journals/jsat/SchubertLB09}, which is necessary for distributed approaches, or through shared memory \cite{manysat}.
In this paper, we transfer these high-level ideas to the area of constraint-driven ILP.
In our approach, the search space is the hypothesis space and workers can co-operate through the exchange of learned constraints, i.e. clauses that describe conflicts.

\section{Problem Setting}
\label{sec:setting}

We now define the LFF problem, on which the approaches in Section \ref{sec:impl} are based.
We assume familiarity with logic programming \cite{lloyd:book}.

The key idea of LFF is to use \emph{hypothesis constraints} to restrict the hypothesis space.
Let $\mathcal{L}$ be a language that defines hypotheses, i.e.~a meta-language.
For instance, consider a meta-language formed of two literals, \emph{h\_lit/4} and \emph{b\_lit/4}, which represent \emph{head} and \emph{body} literals respectively.
With this language, we can denote the clause \emph{last(A,B) $\leftarrow$ tail(A,C), head(C,B)} as the set of literals \emph{\{h\_lit(0,last,2,(0,1)), b\_lit(0,tail,2,(0,2)), b\_lit(0,head,2,(2,1))\}}.
The first argument of each literal is the clause index, the second is the predicate symbol, the third is the arity, and the fourth is the literal variables, where \emph{0} represents \emph{A}, \emph{1} represents \emph{B}, etc.

A \emph{hypothesis constraint} is a constraint expressed in $\mathcal{L}$.
Let $C$ be a set of hypothesis constraints written in a language $\mathcal{L}$.
A set of definite clauses $H$ is \emph{consistent} with $C$ if, when written in $\mathcal{L}$, $H$ does not violate any constraint in $C$.
For instance, the constraint
\emph{$\leftarrow$ h\_lit(0,last,2,(0,1)), b\_lit(0,last,2,(1,0))}
would be violated by the definite clause \emph{last(A,B) $\leftarrow$ last(B,A)}.
Let $\mathcal{H}$ be a hypothesis space.
We denote as $\mathcal{H}_{C}$ the subset of $\mathcal{H}$ which do not violate any constraint in $C$.

% \noindent
We define the LFF problem:

% \begin{definition}[\textbf{LFF input}]
% \label{def:probin}
% The \emph{LFF} problem input is a tuple $(E^+, E^-, B, D, C)$ where
% \begin{itemize}
%     \setlength\itemsep{0pt}
%     \setlength\parskip{0pt}
%     \item $E^+$ and $E^-$ are sets of ground atoms denoting positive and negative examples respectively
%     % \item is a set of ground atoms denoting negative examples
%     \item $B$ is a Horn program denoting background knowledge
%     \item $D$ is a set of predicate declarations
%     \item $C$ is a set of hypothesis constraints
% \end{itemize}
% \end{definition}

\begin{definition}[\textbf{LFF input}]
\label{def:probin}
The \emph{LFF} input is a tuple $(E^+, E^-, B, \mathcal{H}, C)$ where $E^+$ and $E^-$ are sets of ground atoms denoting positive and negative examples respectively; $B$ is a Horn program denoting background knowledge;
$\mathcal{H}$ is a hypothesis space, and $C$ is a set of hypothesis constraints.
% \end{itemize}
\end{definition}

\noindent
% A definite program is a \emph{hypothesis} when it is consistent with both $D$ and $C$.
% We denote the set of such hypotheses as $\mathcal{H}_{D,C}$.
We define a LFF solution:

\begin{definition}[\textbf{LFF solution}]
\label{def:solution}
Given an input tuple $(E^+, E^-, B, \mathcal{H}, C)$, a hypothesis $H \in \mathcal{H}_{C}$ is a \emph{solution} when $H$ is \emph{complete} ($\forall e \in E^+, \; B \cup H \models e$) and \emph{consistent} ($\forall e \in E^-, \; B \cup H \not\models e$).
\end{definition}

\noindent
If a hypothesis is not a solution then it is a \emph{failure} or a \emph{failed hypothesis}.
A hypothesis is \emph{incomplete} when $\exists e \in E^+, \; H \cup B \not \models e$.
A hypothesis is \emph{inconsistent} when $\exists e \in E^-, \; H \cup B \models e$.
A hypothesis is \emph{totally incomplete} when $\forall e \in E^+, \; H \cup B \not \models e$.

Let $cost : \mathcal{H} \mapsto R$ be an arbitrary cost function.
We define an \emph{optimal} solution:

\begin{definition}[\textbf{Optimal solution}]
\label{def:opthyp}
Given an input tuple $(E^+, E^-, B, \mathcal{H}, C)$, a hypothesis $H \in \mathcal{H}_{C}$ is \emph{optimal} when (i) $H$ is a solution, and (ii) $\forall H' \in \mathcal{H}_{C}$, where $H'$ is a solution, $cost(H) \leq cost(H')$.
\end{definition}

\noindent
In this paper, we define the \emph{cost(H)} to be the total number of literals in the logic program $H$.

\subsubsection{Hypothesis Constraints}
% The goal of an LFF learner is to learn hypothesis constraints from failed hypotheses.
\citet{popper,poppi} introduce hypothesis constraints based on subsumption \cite{plotkin:thesis}.
A clause $C_1$ \emph{subsumes} a clause $C_2$ ($C_1 \preceq C_2$) if and only if there exists a substitution $\theta$ such that $C_1\theta \subseteq C_2$.
A clausal theory $T_1$ subsumes a clausal theory $T_2$ ($T_1 \preceq T_2$) if and only if $\forall C_2 \in T_2, \exists C_1 \in T_1$ such that $C_1$ subsumes $C_2$.
A clausal theory $T_1$ is a \emph{specialisation} of a clausal theory $T_2$ if and only if $T_2 \preceq T_1$.
A clausal theory $T_1$ is a \emph{generalisation} of a clausal theory $T_2$ if and only if $T_1 \preceq T_2$.
If a hypothesis $H$ is incomplete, a \emph{specialisation} constraint prunes specialisations of $H$, as they are guaranteed to also be incomplete.
If a hypothesis $H$ is inconsistent, a \emph{generalisation} constraint prunes generalisations of $H$, as they are guaranteed to be inconsistent as well.
If a hypothesis $H$ is totally incomplete, a \emph{redundancy} constraint prunes hypotheses that contain a specialisation of $H$ as a subset.
% \footnote{The full definition is slightly more involved.
% See the work of \citet{poppi} for the full definition.}.
% \ac{something about soundness}

\section{Parallel Algorithms}
\label{sec:impl}

We now describe our parallel ILP approaches.
We first describe the sequential ILP system \name{}, which we parallelise.

\subsection{\name{}}

\noindent
Algorithm \ref{alg:popper} shows the high-level \name{} algorithm, which solves the LFF problem (Definition \ref{def:probin}).
\name{} takes as input positive (\tw{pos}) and negative (\tw{neg}) examples, background knowledge (\tw{bk}), and a maximum hypothesis size (\tw{max\_size}).
\name{} uses  a \emph{generate}, \emph{test}, and \emph{constrain} loop.
\name{} starts with a base \emph{generator} ASP program whose models correspond to hypotheses (definite programs).
% \rolf{As neither $\mathcal{H}$ nor $C$ are arguments, would be good to make the connection in the text, e.g.~including here ``correspond to the hypotheses $\mathcal{H}$''.}
The idea is to augment this generator program with constraints to eliminate models and thus restrict the hypothesis space.
The constraints are initially empty (line 3).

In the generate stage (line 5), \name{} uses Clingo \cite{clingo}, an ASP system, to search for a model of the generator program with exactly \tw{m} literals which \name{} then converts to a hypothesis (a definite program).

In the test stage (line 9), \name{} tests a hypothesis on the given training examples.
If a hypothesis fails, i.e. is \emph{incomplete} or \emph{inconsistent}, then, in the constrain stage (line 12), \name{} learns hypothesis constraints (described as ASP constraints) from the failure.
\name{} adds the constraints to the generator program to prune models and constrain subsequent hypothesis generation.
For instance, if a hypothesis is incomplete, i.e. does not entail all the positive examples, \name{} builds a specialisation constraint to prune hypotheses that are logically more specific.

To find an optimal solution, \name{} progressively increases the number of literals allowed in a hypothesis when the hypothesis space is empty at a certain size (e.g.~when the generator program together with the learned constraints has no more models) (line 6).
% To improve efficiency, \name{} uses Clingo's multi-shot solving \cite{multishot} to maintain state between the three stages and thus remember learned conflicts.
This loop repeats until either (i) \name{} finds an optimal solution, or (ii) there are no more hypotheses to test.
\name{} is guaranteed to find the optimal solution when every hypothesis is guaranteed to terminate, such as when the hypothesis space only contains Datalog programs.

\begin{algorithm}[t]
{
\small
\begin{myalgorithm}[]
def $\text{popper}$(pos, neg, bk, max_size):
  m = 1
  cons = {}
  while m $\leq$ max_size:
    h = generate(cons, m)
    if h == UNSAT:
      m += 1
    else:
      outcome = test(pos, neg, bk, h)
      if outcome == (COMPLETE, CONSISTENT)
        return h
      cons += constrain(h, outcome)
  return UNSAT
\end{myalgorithm}
\caption{
\name{}
}
\label{alg:popper}
}
\end{algorithm}

\subsection{Parallel Solving}
The simplest way to parallelise \name{} is to parallelise the search for a model in the generate stage using the parallel capabilities of Clingo \cite{parclingo}.
% \rolf{Might want to distinguish the notion of ``parallelising the generate stage'' from the most-straightforward implementation of it: using the parallel capabilities of Clingo.}
Clingo incorporates a SAT solver which supports \emph{parallel search} using shared memory multi-threading (cf.~Section \ref{sec:parsat}).
Learned conflict clauses (known as \emph{nogoods}) are exchanged among worker threads according to various heuristics.
To implement this approach, we simply enable Clingo's parallel mode (using the flag `--parallel-mode $k$') which runs $k$ workers in a portfolio configuration.
In our experiments, we call this approach \parname{}.

\subsection{\port{}}
Similar to parallel portfolio SAT approaches, our parallel portfolio ILP approach involves multiple workers searching the same hypothesis space using different strategies.
Algorithm \ref{alg:port} shows the main portfolio worker algorithm, which we call \port{} and which is almost identical to \name{} (Algorithm \ref{alg:popper}).
A master controls the workers.
The master first creates an empty message queue to allow workers to communicate.
This queue is a \emph{many-to-many} queue that allows each worker to receive a copy of any message put on it.
The master then spawns $k$ workers who each search the same hypothesis space.

In addition to the standard \name{} inputs, a \port{} worker receives as input a message queue for communicating constraints (\tw{q\_cons}) and a flag to denote whether to send constraints to other workers.
Each worker calls the generate step (i.e. Clingo) with different search heuristics so that they find models (and thus hypotheses) in a different order to the other workers.
In this paper, we use the simple approach of calling Clingo with the arguments \emph{-{}-rand-freq=p} and \emph{-{}-seed=s}.
The \emph{-{}-rand-freq} flag tells Clingo to perform random (rather than heuristic) decisions with probability $p$.
The value $s$ is a seed.
We set $p=0.01$ and $s$ to be the workerid.
A direction for future work is to determine how to choose a suitable Clingo heuristic.
To be clear, each \port{} worker has its own single-threaded ASP solver.
As with \name{}, \port{} is guaranteed to find the optimal solution when every hypothesis is guaranteed to terminate.

\paragraph{\portc{}}
The \port{} approach, by default, does not permit communication between workers (i.e. the communication flag is false by default).
If the communication option (\tw{com}) is true, then workers exchange constraints with other workers.
We call this communication-enabled version of the algorithm \portc{}.
When one \portc{} worker finds an incomplete or inconsistent hypothesis, it builds the constraints and sends them to the other workers (line 15) and also receives them (line 16).
As \name{} is essentially learning nogoods by testing hypotheses, our sharing of learned constraints can be seen as exchanging \emph{externally learnt} nogoods between multiple solvers.

% \rolf{The normal \port{} strategy is not clearly introduced.
% Coming to the end of this section, it is not clear that the only difference between \port{} and \portc{} is whether \tw{comm} is true and hence whether workers share constraints.}
% \rolf{Might want to consider just using two names:
% Portfolio\textsubscript{com} and Portfolio\textsubscript{\sout{com}}.
% If you go this route, name the procedure in the below algorithm ``Portfolio'' and make the subscript and the argument name consistent (com/comm).}

\begin{algorithm}[t]
{
\small
\begin{myalgorithm}
def $\text{port}$(pos, neg, bk, max_size, q_cons, com):
  m = 1
  cons = {}
  while m $\leq$ max_size:
    h = generate(cons, m)
    if h == UNSAT:
      m += 1
    else:
      outcome = test(pos, neg, bk, h)
      if outcome == (COMPLETE, CONSISTENT)
        return h
      cons' = constrain(h, outcome)
      cons += cons'
      if com:
        q_cons.put(cons')
        cons += q_cons.get()
  return UNSAT
\end{myalgorithm}
\caption{
\port{}
}
\label{alg:port}
}
\end{algorithm}

\subsection{Divide and Conquer}

Similar to parallel D\&C SAT approaches, our parallel D\&C ILP approach involves multiple workers searching disjoint hypothesises spaces.
In this paper, we take the simple approach of splitting the hypothesis space by the hypothesis size.
A direction for future work is to study alternative methods to divide the hypothesis space.

% In other words, each worker searches for a solution with exactly $m$ literals. \rolf{This sentence is ambigious. Could be interpreted as all workers working on size $m$.}

Algorithm \ref{alg:dac} shows the D\&C worker algorithm, which we call \dac{}.
As with \port{}, a master controls the workers.
In addition to creating an empty message queue to share constraints between workers (\tw{q\_cons)}, the master creates a second message queue (\tw{q\_size}) to maintain hypothesis sizes to be searched.
The master populates the size queue with all possible hypothesis sizes in increasing order.
The master then spawns $k$ workers.

In addition to the standard \name{} inputs, a \dac{} worker receives as input a message queue for communicating constraints (\tw{q\_cons}), a message queue for receiving hypothesis sizes to explore (\tw{q\_sizes}), and a flag to denote whether to send constraints to other workers.
A worker pops the smallest size \tw{m} off the  size queue and then searches for a solution with exactly \tw{m} literals.
If there is no solution, the worker loops again and pops another size off the queue.
Since each worker only considers solutions with exactly \tw{m} literals, they all consider disjoint regions of the hypothesis space.
If a worker finds a solution, they inform the master.
If a solution of size $m$ is found, the master waits until all workers searching for hypotheses of size $k < m$ have finished before returning the solution to guarantee that \dac{} returns an optimal solution.

\begin{figure}[t]
\centering
\begin{tikzpicture}

\pgfdeclaredecoration{multiple arrows}{draw}{%
  \state{draw}[width=\pgfdecoratedinputsegmentlength]{%
    \draw [multiple arrows path/.try] (0,0) -- (\pgfdecoratedinputsegmentlength,0);
  }
}
\tikzset{multiple arrows/.style={multiple arrows path/.style={#1},decoration=multiple arrows, decorate}}

\tikzstyle{space}=[thick,ellipse,draw=black,minimum width=100pt,minimum height=50pt,align=center]
%\tikzstyle{circ}=[circle,scale=0.1pt,draw=red]
\tikzstyle{circ}=[scale=0.1pt]
\tikzstyle{goal}=[circle,scale=0.3pt,draw=black,fill=none]

\draw 
%(0,0) node[space] (space1) {}
(5,0) node[space] (space2) {}
%(10,0) node[space] (space3) {};
;
\draw
%($(space1)-(0,1.4)$) node {Sequential}
($(space2)-(0,1.4)$) node {\dac{} (without comm.)}
%($(space3)-(0,1.4)$) node {Portfolio};
;

\draw[dashed] ($(space2)+(-0.8,0.78)$) -- ++(0,-1.56)
($(space2)+(0,0.88)$) -- ++(0,-1.76)
($(space2)+(0.8,0.78)$) -- ++(0,-1.56);
\draw (space2) ++(0.44,0) node {\ldots};
%
%\draw ($(space1)-(0.6,0.6)$) node[style=goal] (G1) {}
\draw ($(space2)-(0.6,0.6)$) node[style=goal] (G2) {};
%($(space3)-(0.6,0.6)$) node[style=goal] (G3) {};
%
%\draw ($(space1)+(0,1.5)$) node (S1S1_) {Searcher}
%(S1S1_.south) node[style=circ] (S1S1) {};

%\draw ($(space2)+(0,1.5)$) node (S2S) {S\textsubscript{1},S\textsubscript{2},\ldots,S$_K$};
%\draw ($(S2S)-(0.875,0.25)$) node[style=circ] (S2S1) {}
%($(S2S)-(0.37,0.25)$) node[style=circ] (S2S2) {}
%($(S2S)+(0.81,-0.25)$) node[style=circ] (S2SK) {};
\draw ($(space2)+(-1.1,2.0)$) node (S2S1) {S$_1$};
\draw ($(space2)+(-0.45,1.2)$) node (S2S2) {S$_2$};
\draw ($(space2)+(0.45,1.2)$) node (S2S3) {S$_3$};
\draw ($(space2)+(1.1,2.0)$) node (S2SK) {S$_k$};

%\draw[thick,densely dotted] (S2S1) -- (S2S2) (S2S1) -- (S2S3) (S2S1) -- (S2SK);
%\draw[thick,densely dotted] (S2S2) -- (S2S3) (S2S2) -- (S2SK);
%\draw[thick,densely dotted] (S2S3) -- (S2SK);
%
%\draw ($(space2)+(-0.05,2.15)$) node[scale=0.55]  {\Letter\hspace{0.5ex}$> >$};
%\draw ($(space2)+(0.05,1.85)$) node[scale=0.55]  {$< <$\hspace{0.5ex}\Letter};

%\draw ($(space3)+(0,1.5)$) node (S3S) {S\textsubscript{1},S\textsubscript{2},\ldots,S$_K$};
%\draw ($(S3S)-(0.875,0.25)$) node[style=circ] (S3S1) {}
%($(S3S)-(0.37,0.25)$) node[style=circ] (S3S2) {}
%($(S3S)+(0.81,-0.25)$) node[style=circ] (S3SK) {};

%\draw[] (S1S1) edge ($(space1)+(0,0.7)$);
%\path [multiple arrows={-latex}] ($(space1)+(0,0.7)$)
%-- ++(180:0.45)
%-- ++(190:0.45)
%-- ++(205:0.45)
%-- ++(240:0.45)
%-- ++(0:0.45)
%-- ++(45:0.45)
%-- ++(15:0.45)
%-- ++(0:0.45)
%-- ++(15:0.45)
%-- ++(-15:0.45)
%-- ++(-30:0.45)
%-- ++(-90:0.45)
%-- ++(140:0.45)
%-- ++(160:0.45)
%-- ++(-70:0.45)
%-- ++(180:0.45)
%-- ++(70:0.45)
%-- ++(200:0.45)
%-- ++(170:0.45)
%-- ++(200:0.45)
%-- ++(215:0.45)
%-- ++(-60:0.45)
%-- (G1)
%;

\draw[] (S2S1) edge ($(space2)+(-0.91,0.59)$);
\def\angle{0}
\path [multiple arrows={-latex}] ($(space2)+(-0.91,0.59)$)
\foreach \stepangle[evaluate=\stepangle as \angle using \angle+\stepangle, remember=\angle] in 
{200,30,120,60,-140,0,210,60}
  { -- ++(\angle:0.45) }
;
\draw[] (S2S2.south)+(0,0.05) edge ($(space2)+(-0.41,0.73)$);
\path [multiple arrows={-stealth}] ($(space2)+(-0.41,0.73)$)
-- ++(-0.35,-0.2)
-- ++(0.7,0)
-- ++(-0.7,-0.3)
-- ++(0.7,0)
-- ++(-0.7,-0.3)
-- ++(0.7,0)
-- ++(-0.7,-0.3)
-- ++(0.7,0)
-- (G2)
;

\draw[] (S2S3.south)+(0,0.05) edge ($(space2)+(0.41,0.73)$);

\draw[] (S2SK) edge ($(space2)+(0.93,0.65)$);
\path [multiple arrows={-Straight Barb}] ($(space2)+(0.93,0.65)$)
\foreach \stepangle[evaluate=\stepangle as \angle using \angle+\stepangle, remember=\angle] in 
{-90,135,-90,-45,-135,90,135,-135}
  { -- ++(\angle:0.45) }
;

%\draw[] (S3S1) edge ($(space3)+(-0.91,0.59)$);
%\path [multiple arrows={-latex}] ($(space3)+(-0.91,0.59)$)
%\foreach \stepangle[evaluate=\stepangle as \angle using \angle+\stepangle, remember=\angle] in 
%{-155,20,70,85,80,-80,-20,-70,-20,55}
%  { -- ++(\angle:0.45) }
%;
%\draw[] (S3S2) edge ($(space3)+(-0.52,0.66)$);
%\path [multiple arrows={-stealth}] ($(space3)+(-0.52,0.66)$)
%\foreach \stepangle[evaluate=\stepangle as \angle using \angle+\stepangle, remember=\angle] in 
%{7,-12,-12,-15,-30,-60,-50,-20,-70}
%  { -- ++(\angle:0.45) }
%;
%
%\draw[] (S3SK) edge ($(space3)+(0.91,0.59)$);
%%\path [multiple arrows={-stealth}] ($(space3)+(0.91,0.59)$)
%%\foreach \stepangle[evaluate=\stepangle as \angle using \angle+\stepangle, remember=\angle] in 
%%{-20,-30,-65,-30,-45,20,-100,20,90,20}
%%  { -- ++(\angle:0.45) }
%%-- (G3)
%%;
%\path [multiple arrows={-Straight Barb}] ($(space3)+(0.91,0.59)$)
%\foreach \stepangle[evaluate=\stepangle as \angle using \angle+\stepangle, remember=\angle] in 
%{-20,-30,-65,-90,-30,30,45,0,20}
%  { -- ++(\angle:0.45) }
%-- (G3)
%;

\end{tikzpicture}
\hspace{2em}
\begin{tikzpicture}
%\usetikzlibrary{arrows}
%\usetikzlibrary{shapes}
%\usetikzlibrary{calc}
%\usetikzlibrary{decorations}
%\usetikzlibrary{arrows.meta}

\pgfdeclaredecoration{multiple arrows}{draw}{%
  \state{draw}[width=\pgfdecoratedinputsegmentlength]{%
    \draw [multiple arrows path/.try] (0,0) -- (\pgfdecoratedinputsegmentlength,0);
  }
}
\tikzset{multiple arrows/.style={multiple arrows path/.style={#1},decoration=multiple arrows, decorate}}

\tikzstyle{space}=[thick,ellipse,draw=black,minimum width=100pt,minimum height=50pt,align=center]
%\tikzstyle{circ}=[circle,scale=0.1pt,draw=red]
\tikzstyle{circ}=[scale=0.1pt]
\tikzstyle{goal}=[circle,scale=0.3pt,draw=black,fill=none]

\draw 
%(0,0) node[space] (space1) {}
(5,0) node[space] (space2) {}
%(10,0) node[space] (space3) {};
;
\draw
%($(space1)-(0,1.4)$) node {Sequential}
($(space2)-(0,1.4)$) node {\dacc{}}
%($(space3)-(0,1.4)$) node {Portfolio};
;

\draw[dashed] ($(space2)+(-0.8,0.78)$) -- ++(0,-1.56)
($(space2)+(0,0.88)$) -- ++(0,-1.76)
($(space2)+(0.8,0.78)$) -- ++(0,-1.56);
\draw (space2) ++(0.44,0) node {\ldots};
%
%\draw ($(space1)-(0.6,0.6)$) node[style=goal] (G1) {}
\draw ($(space2)-(0.6,0.6)$) node[style=goal] (G2) {};
%($(space3)-(0.6,0.6)$) node[style=goal] (G3) {};
%
%\draw ($(space1)+(0,1.5)$) node (S1S1_) {Searcher}
%(S1S1_.south) node[style=circ] (S1S1) {};

%\draw ($(space2)+(0,1.5)$) node (S2S) {S\textsubscript{1},S\textsubscript{2},\ldots,S$_K$};
%\draw ($(S2S)-(0.875,0.25)$) node[style=circ] (S2S1) {}
%($(S2S)-(0.37,0.25)$) node[style=circ] (S2S2) {}
%($(S2S)+(0.81,-0.25)$) node[style=circ] (S2SK) {};
\draw ($(space2)+(-1.1,2.0)$) node (S2S1) {S$_1$};
\draw ($(space2)+(-0.45,1.2)$) node (S2S2) {S$_2$};
\draw ($(space2)+(0.45,1.2)$) node (S2S3) {S$_3$};
\draw ($(space2)+(1.1,2.0)$) node (S2SK) {S$_k$};

\draw[thick,densely dotted] (S2S1) -- (S2S2) (S2S1) -- (S2S3) (S2S1) -- (S2SK);
\draw[thick,densely dotted] (S2S2) -- (S2S3) (S2S2) -- (S2SK);
\draw[thick,densely dotted] (S2S3) -- (S2SK);

\draw ($(space2)+(-0.05,2.15)$) node[scale=0.55]  {\Letter\hspace{0.5ex}$> >$};
\draw ($(space2)+(0.05,1.85)$) node[scale=0.55]  {$< <$\hspace{0.5ex}\Letter};

%\draw ($(space3)+(0,1.5)$) node (S3S) {S\textsubscript{1},S\textsubscript{2},\ldots,S$_K$};
%\draw ($(S3S)-(0.875,0.25)$) node[style=circ] (S3S1) {}
%($(S3S)-(0.37,0.25)$) node[style=circ] (S3S2) {}
%($(S3S)+(0.81,-0.25)$) node[style=circ] (S3SK) {};

%\draw[] (S1S1) edge ($(space1)+(0,0.7)$);
%\path [multiple arrows={-latex}] ($(space1)+(0,0.7)$)
%-- ++(180:0.45)
%-- ++(190:0.45)
%-- ++(205:0.45)
%-- ++(240:0.45)
%-- ++(0:0.45)
%-- ++(45:0.45)
%-- ++(15:0.45)
%-- ++(0:0.45)
%-- ++(15:0.45)
%-- ++(-15:0.45)
%-- ++(-30:0.45)
%-- ++(-90:0.45)
%-- ++(140:0.45)
%-- ++(160:0.45)
%-- ++(-70:0.45)
%-- ++(180:0.45)
%-- ++(70:0.45)
%-- ++(200:0.45)
%-- ++(170:0.45)
%-- ++(200:0.45)
%-- ++(215:0.45)
%-- ++(-60:0.45)
%-- (G1)
%;

\draw[] (S2S1) edge ($(space2)+(-0.91,0.59)$);
\def\angle{0}
\path [multiple arrows={-latex}] ($(space2)+(-0.91,0.59)$)
\foreach \stepangle[evaluate=\stepangle as \angle using \angle+\stepangle, remember=\angle] in 
%{200,30,120,60,-140,0,210,60}
{200,25,130,-35,-115,-60}
  { -- ++(\angle:0.45) }
;

\draw[] (S2S2.south)+(0,0.05) edge ($(space2)+(-0.41,0.73)$);
\path [multiple arrows={-stealth}] ($(space2)+(-0.41,0.73)$)
-- ++(-0.35,-0.2)
-- ++(0.7,0)
%%-- ++(-0.7,-0.3)
-- ++(0,-0.3)
%-- ++(0.7,0)
%%-- ++(-0.7,-0.3)
%-- ++(0,-0.3)
-- ++(-0.7,0)
-- ++(0.7,-0.3)
-- ++(-0.7,-0.3)
-- ++(0.7,0)
-- (G2)
;

\draw[] (S2S3.south)+(0,0.05) edge ($(space2)+(0.41,0.73)$);

\draw[] (S2SK) edge ($(space2)+(0.93,0.65)$);
\path [multiple arrows={-Straight Barb}] ($(space2)+(0.93,0.65)$)
\foreach \stepangle[evaluate=\stepangle as \angle using \angle+\stepangle, remember=\angle] in 
{-90,135,-135,-45,135,-135}
%-90,-45,-135,90,135,-135}
  { -- ++(\angle:0.45) }
;

%\draw[] (S3S1) edge ($(space3)+(-0.91,0.59)$);
%\path [multiple arrows={-latex}] ($(space3)+(-0.91,0.59)$)
%\foreach \stepangle[evaluate=\stepangle as \angle using \angle+\stepangle, remember=\angle] in 
%{-155,20,70,85,80,-80,-20,-70,-20,55}
%  { -- ++(\angle:0.45) }
%;
%\draw[] (S3S2) edge ($(space3)+(-0.52,0.66)$);
%\path [multiple arrows={-stealth}] ($(space3)+(-0.52,0.66)$)
%\foreach \stepangle[evaluate=\stepangle as \angle using \angle+\stepangle, remember=\angle] in 
%{7,-12,-12,-15,-30,-60,-50,-20,-70}
%  { -- ++(\angle:0.45) }
%;
%
%\draw[] (S3SK) edge ($(space3)+(0.91,0.59)$);
%%\path [multiple arrows={-stealth}] ($(space3)+(0.91,0.59)$)
%%\foreach \stepangle[evaluate=\stepangle as \angle using \angle+\stepangle, remember=\angle] in 
%%{-20,-30,-65,-30,-45,20,-100,20,90,20}
%%  { -- ++(\angle:0.45) }
%%-- (G3)
%%;
%\path [multiple arrows={-Straight Barb}] ($(space3)+(0.91,0.59)$)
%\foreach \stepangle[evaluate=\stepangle as \angle using \angle+\stepangle, remember=\angle] in 
%{-20,-30,-65,-90,-30,30,45,0,20}
%  { -- ++(\angle:0.45) }
%-- (G3)
%;

\end{tikzpicture}
\caption{Divide-and-conquer strategy with and without communication.
With communication enabled, learned constraints are passed as messages between workers.}
\label{fig:communication}
\end{figure}
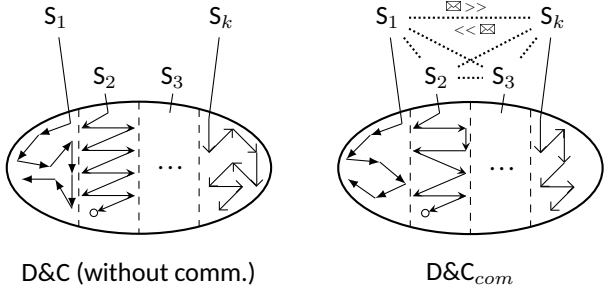

\paragraph{\dacc{}}
The \dac{} approach, by default, does not permit communication between workers (i.e. the communication is defaulted to false).
If the communication option (\tw{com}) is true, then workers exchange constraints with other workers.
We call this communication-enabled version of the algorithm \dacc{}.
Figure \ref{fig:communication} illustrates the difference between \dac{} and \dacc{}.

\begin{algorithm}[t]
{
% \small
\begin{myalgorithm}[]
def $\text{dac}$(pos, neg, bk, q_size, q_cons, com):
  cons = {}
  while |q_size| > 1:
    m = q_size.get()
    while true:
      h = generate(cons, m)
      if h == UNSAT:
        break
      outcome = test(pos, neg, bk, h)
      if outcome == (COMPLETE, CONSISTENT)
        return h
      cons' = constrain(h, outcome)
      cons += cons'
      if com:
        q_cons.put(cons')
        cons += q_cons.get()
  return UNSAT
\end{myalgorithm}
\caption{
\dac{}
}
\label{alg:dac}
}
\end{algorithm}

\subsection{Implementation}
% \ac{@JK, do you want to write this section?}
We implemented the systems in Python 3 using the multiprocessing module.
Each process is an instance of Popper which maintains its own ASP solver.
Rather than use shared memory, we use a message queue to allow workers to communicate.
% , partly to support future distribute approaches.

% program generator (solver), constraint grounder, and hypothesis tester.
% The solver and grounder both use a Clingo backend \ref{clingo:guide}, and the tester uses a Prolog \ref{DBLP:journals/sigart/WarrenP077} backend.

\section{Experiments}
\label{sec:exp}

We claim that our parallelisation approaches can reduce learning times and thus improve scalability.
Our experiments\footnote{
All the implementation code and experimental data will be made open-source and freely available after the paper has gone through review.
} therefore aim to answer the question:

\begin{description}
\item[Q1] Can our parallel approaches reduce learning times?
\end{description}

\noindent
To answer this question, we compare the performance of our approaches when given progressively more workers (CPUs).

Our parallel approaches allow workers to exchange learned constraints.
To evaluate the impact of communication, our experiments aim to answer the question:

\begin{description}
\item[Q2] Can communication (sharing constraints) reduce learning times?
\end{description}

\paragraph{Settings.}
We use Clingo version 5.5.0, SWI-Prolog 7.6.3 and Python 3.9.6.
% \rolf{We should probably list the version of clingo we use, especially because of ParSearch and the options we specify in the previous section.}
All experiments were performed on a server with an AMD Opteron\textsuperscript{TM} Processor with 32 cores, 64 threads, and 256GB of RAM.
% \rolf{As a reviewer, I would like to know why not more cores are used in the experiments.}
We give all the approaches identical inputs in all the experiments.
% We measure mean learning times.
We enforce a timeout of five minutes per task.
We repeat each experiment five times and measure mean learning time and standard deviation.
% Because the representation language (definite programs) is Turing-complete, \name{} can generate non-terminating programs.
% We therefore enforce a testing timeout of 0.01 seconds per example.
% If a program times out, we view it as a failure.
% \noindent

\subsection{Domains}

We consider two domains: \emph{program synthesis} and \emph{inductive general game playing}.

\subsubsection{Program synthesis}

Program synthesis has long been considered a difficult problem in ILP \cite{ilp20}.
Indeed, most ILP systems cannot learn recursive programs.
\citet{popper} showed that \name{} can learn recursive programs with higher accuracies and in less time than other ILP systems.
In this experiment, we try to answer the experimental questions using a similar program synthesis dataset.

We use four challenging synthesis tasks: (\emph{find dupl}) find a duplicate element in a list,  (\emph{sorted}) determine whether a list is sorted, (\emph{dropk}) drop the first k elements in a list, and (\emph{filter}) remove all odd elements from a list.
Figure \ref{fig:filter} shows an example solution for the \emph{filter} task.
We provide as background knowledge the dyadic relations \emph{head}, \emph{tail}, \emph{element}, \emph{increment}, \emph{decrement}, \emph{geq} and the monadic relations \emph{empty}, \emph{zero}, \emph{one}, \emph{even}, and \emph{odd}.
We also include the triadic relation \emph{prepend} in the \emph{filter} experiment.
% The exact settings can be found in Appendix \ref{app:lists}.
% \rolf{reference to appendix broken on my end}

\begin{figure}
\begin{code}
f(A,B):- empty(A),empty(B).\\
f(A,B):- head(A,D),odd(D),tail(A,C),f(C,B).\\
f(A,B):- tail(A,C),head(A,E),even(E),\\
\quad\quad\quad\quad f(C,D),prepend(E,D,B).\\
\end{code}
\caption{An example solution described as Prolog program for the \emph{filter} task.}
\label{fig:filter}
\end{figure}

For each task, we generate 10 positive and 10 negative training examples.
Each list is randomly generated and has a maximum length of 50.
We sample the list elements uniformly at random from the set $\{1,2,\dots,100\}$.

\subsubsection{IGGP}
The general game playing (GGP) framework~\cite{ggp} is a system for evaluating an agent's general intelligence across a wide range of tasks.
In the GGP competition, agents are tested on games they have never seen before.
In each round, the agents are given the rules of a new game. The rules are described symbolically as a logic program.
The agents are given a few seconds to process the rules of the game, then they start playing, thus producing game traces.
The winner of the competition is the agent who gets the best total score over all the games.
In this experiment, we use the IGGP dataset \cite{iggp} which inverts the GGP task: an ILP system is given game traces and the task is to learn a set of rules (a logic program) that could have produced these traces.
% As \citet{iggp} show, the IGGP dataset is extremely difficult for ILP systems.
We focus on two IGGP games: \emph{minimal decay} and \emph{rock, paper, scissors} (rps) and on learning the \emph{next} relations, which is the most challenging one to learn \cite{iggp}.
Figure \ref{fig:rps} shows an example solution for the \emph{rps} task.

\begin{figure}
\begin{code}
nextscore(A,B,C):- score(A,B,C),does(A,D,F),\\
\quad\quad\quad\quad\quad does(A,E,F),different(D,E).\\
nextscore(A,B,C):- different(B,E),does(A,E,G),\\
\quad\quad\quad\quad\quad score(A,F,C),does(A,F,D),beats(G,D).\\
nextscore(A,B,C):- score(A,B,F),beats(E,G),\\
\quad\quad\quad\quad\quad does(A,D,G),succ(F,C),\\
\quad\quad\quad\quad\quad different(B,D),does(A,B,E).\\
\end{code}
\caption{An example solution described as Prolog program for the \emph{rps} task.}
\label{fig:rps}
\end{figure}

\subsection{Results}

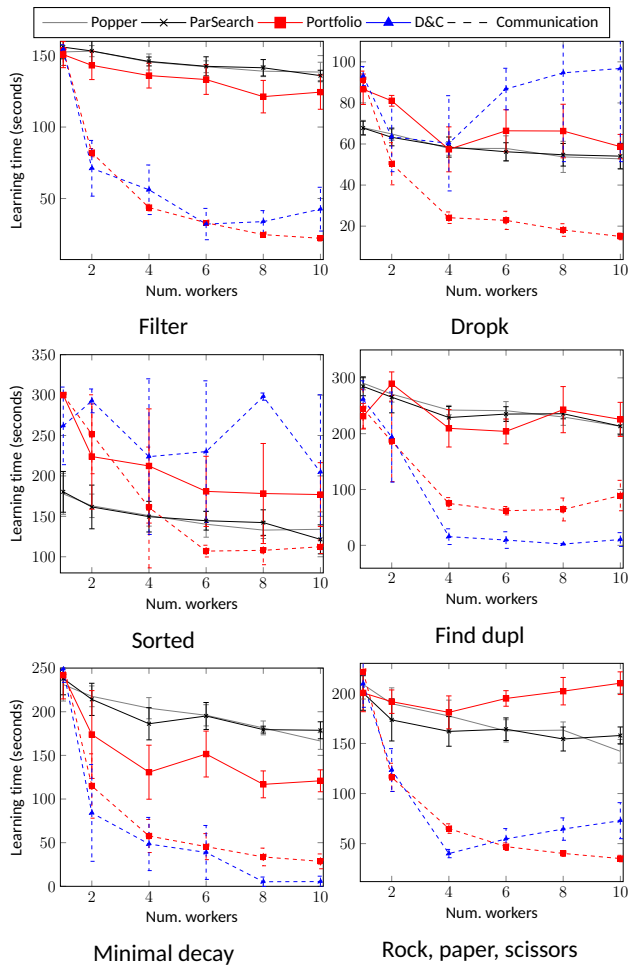
\begin{figure}[ht]
\centering
\begin{tikzpicture}
\begin{customlegend}[legend columns=5,legend style={nodes={scale=.6, transform shape},align=left,column sep=0ex},
        legend entries={Popper, \parname{}, \port{}, \dac{}, Communication}]
        \addlegendimage{mark=none,gray}
        \addlegendimage{mark=x,black}
        \addlegendimage{mark=square*,red}
        \addlegendimage{mark=triangle*,blue}
        \addlegendimage{dashed}
        \end{customlegend}
\end{tikzpicture}
\begin{subfigure}{.5\linewidth}
\centering
\pgfplotsset{scaled x ticks=false}
\pgfplotsset{every tick label/.append style={font=\Large}}
\begin{tikzpicture}[scale=.51]
    \begin{axis}[
    xlabel=Num. workers,
    ylabel=Learning time (seconds),
    xmin=0.9,
    xmax=10.1,
    ymax=160,
    ylabel style={yshift=-1mm},
    label style={font=\Large},
    legend style={legend pos=south west,font=\scriptsize,style={nodes={right}}}
    ]

% standard popper
\addplot+[gray,mark=none,error bars/.cd,y dir=both,y explicit]
table [
x=workers,
y=time,
col sep=comma,
y error plus expr=\thisrow{error},
y error minus expr=\thisrow{error},
] {data/filter-default_popper.csv};

% parallel clingo
\addplot+[black,mark=x,error bars/.cd,y dir=both,y explicit]
table [
x=workers,
y=time,
col sep=comma,
y error plus expr=\thisrow{error},
y error minus expr=\thisrow{error},
] {data/filter-parallel_clingo.csv};

% portfolio
\addplot+[red,mark=square*,mark options={fill=red},error bars/.cd,y dir=both,y explicit]
table [
x=workers,
y=time,
col sep=comma,
y error plus expr=\thisrow{error},
y error minus expr=\thisrow{error},
] {data/filter-portfolio.csv};

% % portfolio-comm
\addplot+[red,mark=square*,dashed,mark options={fill=red},error bars/.cd,y dir=both,y explicit]
table [
x=workers,
y=time,
col sep=comma,
y error plus expr=\thisrow{error},
y error minus expr=\thisrow{error},
] {data/filter-portfolio_comm.csv};

% % dac
% \addplot+[blue,mark=triangle*,mark options={fill=blue},error bars/.cd,y dir=both,y explicit]
% table [
% x=workers,
% y=time,
% col sep=comma,
% y error plus expr=\thisrow{error},
% y error minus expr=\thisrow{error},
% ] {data/filter-dac.csv};

% dac-comm
\addplot+[blue,mark=triangle*,dashed,mark options={fill=blue},error bars/.cd,y dir=both,y explicit]
table [
x=workers,
y=time,
col sep=comma,
y error plus expr=\thisrow{error},
y error minus expr=\thisrow{error},
] {data/filter-dac_comm.csv};

    \end{axis}
  \end{tikzpicture}
\caption*{Filter}
% \label{fig:filter}
\end{subfigure}%
\begin{subfigure}{.5\linewidth}
\centering
\pgfplotsset{scaled x ticks=false}
\pgfplotsset{every tick label/.append style={font=\Large}}
\begin{tikzpicture}[scale=.51]
    \begin{axis}[
    xlabel=Num. workers,
    % ylabel=Learning time (seconds),
    xmin=0.9,
    xmax=10.1,
    ymax=110,
    ylabel style={yshift=-1mm},
    label style={font=\Large},
    legend style={legend pos=south west,font=\scriptsize,style={nodes={right}}}
    ]

% standard popper
\addplot+[gray,mark=none,error bars/.cd,y dir=both,y explicit]
table [
x=workers,
y=time,
col sep=comma,
y error plus expr=\thisrow{error},
y error minus expr=\thisrow{error},
] {data/dropk-default_popper.csv};

% parallel clingo
\addplot+[black,mark=x,error bars/.cd,y dir=both,y explicit]
table [
x=workers,
y=time,
col sep=comma,
y error plus expr=\thisrow{error},
y error minus expr=\thisrow{error},
] {data/dropk-parallel_clingo.csv};

% portfolio
\addplot+[red,mark=square*,mark options={fill=red},error bars/.cd,y dir=both,y explicit]
table [
x=workers,
y=time,
col sep=comma,
y error plus expr=\thisrow{error},
y error minus expr=\thisrow{error},
] {data/dropk-portfolio.csv};

% % portfolio-comm
\addplot+[red,mark=square*,dashed,mark options={fill=red},error bars/.cd,y dir=both,y explicit]
table [
x=workers,
y=time,
col sep=comma,
y error plus expr=\thisrow{error},
y error minus expr=\thisrow{error},
] {data/dropk-portfolio_comm.csv};

% % % dac
% \addplot+[blue,mark=triangle*,mark options={fill=blue},error bars/.cd,y dir=both,y explicit]
% table [
% x=workers,
% y=time,
% col sep=comma,
% y error plus expr=\thisrow{error},
% y error minus expr=\thisrow{error},
% ] {data/dropk-dac.csv};

% % % dac-comm
\addplot+[blue,mark=triangle*,dashed,mark options={fill=blue},error bars/.cd,y dir=both,y explicit]
table [
x=workers,
y=time,
col sep=comma,
y error plus expr=\thisrow{error},
y error minus expr=\thisrow{error},
] {data/dropk-dac_comm.csv};

    \end{axis}
  \end{tikzpicture}
  \caption*{Dropk}
\end{subfigure} %

\begin{subfigure}{.5\linewidth}
\centering
\pgfplotsset{scaled x ticks=false}
\pgfplotsset{every tick label/.append style={font=\Large}}
\begin{tikzpicture}[scale=.51]
    \begin{axis}[
    xlabel=Num. workers,
    ylabel=Learning time (seconds),
    xmin=0.9,
    xmax=10.1,
    ymin=80,
    ymax=350,
    ylabel style={yshift=-1mm},
    label style={font=\Large},
    legend style={legend pos=south west,font=\scriptsize,style={nodes={right}}}
    ]

% % standard popper
\addplot+[gray,mark=none,error bars/.cd,y dir=both,y explicit]
table [
x=workers,
y=time,
col sep=comma,
y error plus expr=\thisrow{error},
y error minus expr=\thisrow{error},
] {data/sorted-default_popper.csv};

% % parallel clingo
\addplot+[black,mark=x,error bars/.cd,y dir=both,y explicit]
table [
x=workers,
y=time,
col sep=comma,
y error plus expr=\thisrow{error},
y error minus expr=\thisrow{error},
] {data/sorted-parallel_clingo.csv};

% % portfolio
\addplot+[red,mark=square*,mark options={fill=red},error bars/.cd,y dir=both,y explicit]
table [
x=workers,
y=time,
col sep=comma,
y error plus expr=\thisrow{error},
y error minus expr=\thisrow{error},
] {data/sorted-portfolio.csv};

% % portfolio-comm
\addplot+[red,mark=square*,dashed,mark options={fill=red},error bars/.cd,y dir=both,y explicit]
table [
x=workers,
y=time,
col sep=comma,
y error plus expr=\thisrow{error},
y error minus expr=\thisrow{error},
] {data/sorted-portfolio_comm.csv};

% % dac
% \addplot+[blue,mark=triangle*,mark options={fill=blue},error bars/.cd,y dir=both,y explicit]
% table [
% x=workers,
% y=time,
% col sep=comma,
% y error plus expr=\thisrow{error},
% y error minus expr=\thisrow{error},
% ] {data/sorted-dac.csv};

% % % dac-comm
\addplot+[blue,mark=triangle*,dashed,mark options={fill=blue},error bars/.cd,y dir=both,y explicit]
table [
x=workers,
y=time,
col sep=comma,
y error plus expr=\thisrow{error},
y error minus expr=\thisrow{error},
] {data/sorted-dac_comm.csv};

    \end{axis}
  \end{tikzpicture}
\caption*{Sorted}
% \label{fig:filter}
\end{subfigure}%
\begin{subfigure}{.5\linewidth}
\centering
\pgfplotsset{scaled x ticks=false}
\pgfplotsset{every tick label/.append style={font=\Large}}
\begin{tikzpicture}[scale=.51]
    \begin{axis}[
    xlabel=Num. workers,
    % ylabel=Learning time (seconds),
    xmin=0.9,
    xmax=10.1,
    ymax=350,
    ylabel style={yshift=-1mm},
    label style={font=\Large},
    legend style={legend pos=south west,font=\scriptsize,style={nodes={right}}}
    ]

% % standard popper
\addplot+[gray,mark=none,error bars/.cd,y dir=both,y explicit]
table [
x=workers,
y=time,
col sep=comma,
y error plus expr=\thisrow{error},
y error minus expr=\thisrow{error},
] {data/find-dupl-default_popper.csv};

% % parallel clingo
\addplot+[black,mark=x,error bars/.cd,y dir=both,y explicit]
table [
x=workers,
y=time,
col sep=comma,
y error plus expr=\thisrow{error},
y error minus expr=\thisrow{error},
] {data/find-dupl-parallel_clingo.csv};

% % portfolio
\addplot+[red,mark=square*,mark options={fill=red},error bars/.cd,y dir=both,y explicit]
table [
x=workers,
y=time,
col sep=comma,
y error plus expr=\thisrow{error},
y error minus expr=\thisrow{error},
] {data/find-dupl-portfolio.csv};

% % portfolio-comm
\addplot+[red,mark=square*,dashed,mark options={fill=red},error bars/.cd,y dir=both,y explicit]
table [
x=workers,
y=time,
col sep=comma,
y error plus expr=\thisrow{error},
y error minus expr=\thisrow{error},
] {data/find-dupl-portfolio_comm.csv};

% dac
% \addplot+[blue,mark=triangle*,mark options={fill=blue},error bars/.cd,y dir=both,y explicit]
% table [
% x=workers,
% y=time,
% col sep=comma,
% y error plus expr=\thisrow{error},
% y error minus expr=\thisrow{error},
% ] {data/find-dupl-dac.csv};

% % dac-comm
\addplot+[blue,mark=triangle*,dashed,mark options={fill=blue},error bars/.cd,y dir=both,y explicit]
table [
x=workers,
y=time,
col sep=comma,
y error plus expr=\thisrow{error},
y error minus expr=\thisrow{error},
] {data/find-dupl-dac_comm.csv};

    \end{axis}
  \end{tikzpicture}
  \caption*{Find dupl}
\end{subfigure} %

\begin{subfigure}{.5\linewidth}
\centering
\pgfplotsset{scaled x ticks=false}
\pgfplotsset{every tick label/.append style={font=\Large}}
\begin{tikzpicture}[scale=.51]
    \begin{axis}[
    xlabel=Num. workers,
    ylabel=Learning time (seconds),
    xmin=0.9,
    xmax=10.1,
    ymin=0,
    ymax=250,
    ylabel style={yshift=-1mm},
    label style={font=\Large},
    legend style={legend pos=south west,font=\scriptsize,style={nodes={right}}}
    ]

% standard popper
\addplot+[gray,mark=none,error bars/.cd,y dir=both,y explicit]
table [
x=workers,
y=time,
col sep=comma,
y error plus expr=\thisrow{error},
y error minus expr=\thisrow{error},
] {data/iggp-minimal-decay-default_popper.csv};

% parallel clingo
\addplot+[black,mark=x,error bars/.cd,y dir=both,y explicit]
table [
x=workers,
y=time,
col sep=comma,
y error plus expr=\thisrow{error},
y error minus expr=\thisrow{error},
] {data/iggp-minimal-decay-parallel_clingo.csv};

% portfolio
\addplot+[red,mark=square*,mark options={fill=red},error bars/.cd,y dir=both,y explicit]
table [
x=workers,
y=time,
col sep=comma,
y error plus expr=\thisrow{error},
y error minus expr=\thisrow{error},
] {data/iggp-minimal-decay-portfolio.csv};

% % portfolio-comm
\addplot+[red,mark=square*,dashed,mark options={fill=red},error bars/.cd,y dir=both,y explicit]
table [
x=workers,
y=time,
col sep=comma,
y error plus expr=\thisrow{error},
y error minus expr=\thisrow{error},
] {data/iggp-minimal-decay-portfolio_comm.csv};

% % dac
% \addplot+[blue,mark=triangle*,mark options={fill=blue},error bars/.cd,y dir=both,y explicit]
% table [
% x=workers,
% y=time,
% col sep=comma,
% y error plus expr=\thisrow{error},
% y error minus expr=\thisrow{error},
% ] {data/iggp-minimal-decay-dac.csv};

% dac-comm
\addplot+[blue,mark=triangle*,dashed,mark options={fill=blue},error bars/.cd,y dir=both,y explicit]
table [
x=workers,
y=time,
col sep=comma,
y error plus expr=\thisrow{error},
y error minus expr=\thisrow{error},
] {data/iggp-minimal-decay-dac_comm.csv};

    \end{axis}
  \end{tikzpicture}
\caption*{Minimal decay}
% \label{fig:iggp-minimal-decay}
\end{subfigure}%
\begin{subfigure}{.5\linewidth}
\centering
\pgfplotsset{scaled x ticks=false}
\pgfplotsset{every tick label/.append style={font=\Large}}
\begin{tikzpicture}[scale=.51]
    \begin{axis}[
    xlabel=Num. workers,
    % ylabel=Learning time (seconds),
    xmin=0.9,
    xmax=10.1,
    ymax=230,
    ylabel style={yshift=-1mm},
    label style={font=\Large},
    legend style={legend pos=south west,font=\scriptsize,style={nodes={right}}}
    ]

% % standard popper
\addplot+[gray,mark=none,error bars/.cd,y dir=both,y explicit]
table [
x=workers,
y=time,
col sep=comma,
y error plus expr=\thisrow{error},
y error minus expr=\thisrow{error},
] {data/iggp-rps-default_popper.csv};

% parallel clingo
\addplot+[black,mark=x,error bars/.cd,y dir=both,y explicit]
table [
x=workers,
y=time,
col sep=comma,
y error plus expr=\thisrow{error},
y error minus expr=\thisrow{error},
] {data/iggp-rps-parallel_clingo.csv};

% portfolio
\addplot+[red,mark=square*,mark options={fill=red},error bars/.cd,y dir=both,y explicit]
table [
x=workers,
y=time,
col sep=comma,
y error plus expr=\thisrow{error},
y error minus expr=\thisrow{error},
] {data/iggp-rps-portfolio.csv};

% % portfolio-comm
\addplot+[red,mark=square*,dashed,mark options={fill=red},error bars/.cd,y dir=both,y explicit]
table [
x=workers,
y=time,
col sep=comma,
y error plus expr=\thisrow{error},
y error minus expr=\thisrow{error},
] {data/iggp-rps-portfolio_comm.csv};

% % dac
% \addplot+[blue,mark=triangle*,mark options={fill=blue},error bars/.cd,y dir=both,y explicit]
% table [
% x=workers,
% y=time,
% col sep=comma,
% y error plus expr=\thisrow{error},
% y error minus expr=\thisrow{error},
% ] {data/iggp-rps-dac.csv};

% dac-comm
\addplot+[blue,mark=triangle*,dashed,mark options={fill=blue},error bars/.cd,y dir=both,y explicit]
table [
x=workers,
y=time,
col sep=comma,
y error plus expr=\thisrow{error},
y error minus expr=\thisrow{error},
] {data/iggp-rps-dac_comm.csv};

    \end{axis}
  \end{tikzpicture}
\caption*{Rock, paper, scissors}
\end{subfigure}

\caption{Experimental results.}
\label{fig:results}
\end{figure}

Figure \ref{fig:results} shows the experimental results.
\name{} is the single-core baseline.

\paragraph{\parname{}.}
The results show no major benefit from \parname{}, i.e. running \name{} with parallel Clingo enabled.
For instance, in the \emph{filter} experiment, the difference in running time between \name{} and \parname{} is less than 5\%.
This result may surprise the reader, especially as parallel Clingo has been shown to outperform sequential Clingo \cite{parclingo}.
The reason is that the bottleneck in \name{} is rarely generating a hypothesis, i.e. searching for a syntactically valid program.
Instead, the bottleneck is mostly the sheer number of hypotheses to consider.
There is, therefore, little benefit from parallelising \emph{only} the generate part.
% \ac{maybe something about showing unsat?}

\paragraph{\port{}.}
The results are mixed for the \port{} approach without communication.
In two tasks it performs worse than \name{} (\emph{dropk} and \emph{sorted}), in one better (\emph{minimal decay}), and about the same in the rest.
This result is expected.
The only way that \port{} can outperform \name{} is when a worker happens to find a solution quicker than \name{} because of its different search heuristics.
However, since \name{} performs iterative deepening search on the hypothesis sizes, whereby it proves that there is no solution at a certain size before going to the next size, both \name{} and \port{} should roughly take the same amount of time to search the space of programs smaller than the solution.
It is only when searching the part of the space at the size of the solution that \port{} has the opportunity to outperform \name{}, hence the modest improvements.

\paragraph{\portc{}.}
The results are very strong for the \portc{} approach.
In all cases, \portc{} outperforms \name{} and a paired t-test confirms the significance at the $p < 0.01$ level.
For instance, in the \emph{minimal decay} experiment, given one worker, \portc{} takes about the same time (around 250s) as \name{}, which is to be expected as there is no parallelisation with only one worker.
Given more workers, \portc{} starts to outperform \name{}.
With two workers, the learning time of \portc{} is almost halved to 136s.
With four workers, the learning time of \portc{} is halved again to 60s.
With eight workers, the learning time of \portc{} is almost halved again to around 35s.
Similar reductions are demonstrated in all the tasks.
This result strongly suggests that the answer to \textbf{Q1} is yes: our parallelisation approaches can significantly reduce learning times and that the speed-up is roughly proportional to the number of workers.
In all cases, \portc{} outperforms \port{} and a paired t-test confirms the significance at the $p < 0.01$ level.
This result clearly demonstrates that communication between workers is important for good learning performance (\textbf{Q2}).

\paragraph{\dac{}.}
Figure \ref{fig:results} excludes the results of \dac{} (\dac{} without communication) results.
We have excluded the results for \dac{} because it struggles to find solutions for any tasks in the given time limit.
This result may surprise the reader, as they may think that \dac{} would in the worst-case simulate \popper{}.
However, the key omission of \dac{} is the inability to learn constraints from failed hypotheses from multiple hypothesis sizes.
A key reason why \popper{} can efficiently find solutions is that it first considers smaller hypotheses before larger ones.
By learning constraints from small failed hypotheses, \popper{} can prune large parts of the hypothesis space.
% \ac{give an exampke}.
So when \popper{} increases the hypothesis size bound, the space for the next size is greatly reduced from constraints learnt when learning for smaller sizes.
However, in the \dac{} approach, there is no transfer of knowledge between the hypothesis sizes.
By contrast, \dacc{} performs reasonably well, outperforming \popper{} on four of the six tasks.
The most impressive results are in the \emph{minimal decay} task.
With one worker, \dacc{} takes around 250s.
With two workers, the running time of \dacc{} is reduced to one fifth (45s).
Given eight workers, \dacc{} takes around 1s to find a solution.
% \rolf{Impressive indeed. Without an explanation I would be a bit suspicious as a reviewer...}
This result again clearly demonstrates that communication between workers is important for good learning performance (\textbf{Q2}).

\section{Conclusions and Limitations}

To improve the ILP scalability problem, we have introduced parallel ILP approaches inspired by parallel SAT approaches, namely \emph{portfolio} and \emph{divide-and-conquer} approaches.
Our implementations parallelise \name{}, a state-of-the-art constraint-driven ILP system.
Our experiments on two domains (program synthesis and inductive general game playing) show that (i) our parallel methods can lead to linear speedups with up to four processors in general, (ii) our parallel methods can lead to super-linear speedup in some cases, and (iii) that communication (i.e. sharing constraints) is important for good performance.
As far as we are aware, this work is the first to clearly demonstrate the ability to parallelise state-of-the-art ILP systems that support predicate invention and learning recursive programs.

\subsection{Limitations and Future Work}

There are many limitations of this work, which open much scope for future work.

\paragraph{Better implementation.}
Our implementations are based on message passing via inter-process communication.
As our workers learn many, and at times large, constraints, message passing involves non-trivial overhead, not least due to copying.
On a single machine, we could exploit shared memory for improved performance, e.g.~due to less copying.

%\rolf{A better implementation based on shared-memory instead of inter-process communication.
%Not having to copy the many, and sometimes large, (ground) constraints just for communication purposed should help.
%This would correspond with the performance improvement parallel SAT solver got from going from distributed approaches to single machine shared-memory implementations.}

\paragraph{Better constraints.}
Parallel SAT solver workers tend to perform checks on the clauses that they receive.
One such check ensures that a new \emph{propositional} constraint is not subsumed by any known constraint \cite{manysat}.
We could also check subsumption of our learned \emph{first-order} constraints, though this does have non-trivial cost.
SAT solvers also employ heuristics for deciding which constraints to share.
For example, typically only small learned constraints are sent to other workers, based on some cut-off value \cite{parsat}.
Another heuristic is to only send (long) constraints to workers searching a similar part of the solution space (ibid.).
We could empirically test if parallel SAT techniques like these are beneficial to parallel constraint-driven ILP.

%\rolf{SAT solver workers do some checks on shared clauses.
%One of them is that no subsumed clauses are added.
%We could do something similar, but at first-order, which is more costly.}
%\rolf{SAT solvers typically use a cut-off value of sizes of learnt clauses to decide whether to share or not.}
%\rolf{
%SAT literature talks about strategies for only sharing clauses with workers working on a similar part of the search space.
%We could argue that we want such a cut-off as otherwise you're sharing constraints for part of the search space that the other searchers will never reach.}
%\ac{@RM can you please polish the two points above?}

\paragraph{Distributed approaches.}
Our implementations support local parallelisation, i.e.~on one machine.
Our choice for message passaging is an advantage here, as it should generalise to distributed parallelisation, where we could potentially harness hundreds of CPUs.
% \ac{However, having more workers can potentially lead to much duplicated search ... }

% \paragraph{Better heuristics.}
% Most parallel SAT solvers seem to rely on parameter tuning to get complementary competing solver instances.
% We have left this unexplored.

\paragraph{Combination of \dac{} and \port{}}
The D\&C approach sometimes has one worker working on size $k$ and other workers doing nothing, as they have finished working on other sizes.
Instead of them waiting, they could start work on the same size $k$, each with a new search heuristic.
In line with a distributed approach having many workers, we could follow the parallel SAT strategy of first splitting the search space and then applying a portfolio of workers to each subspace.

%\rolf{These are hybrid approaches in the SAT literature.
%Typically they first split a search space and then put a portfolio of workers on each subspace.
%}
%\ac{@RM can you please polish the point above}

% \paragraph{Share nogoods?}
% \ac{Anyone?}

% Improved Knowledge Sharing

% To alleviate these problems, Hamadi, Jabbour, and Sais (2009) have introduced a dynamic strate- gy that uses control theory techniques to increase or reduce automatically the quantity of clauses shared between two search efforts.

% \bibliographystyle{aaai}
\bibliography{mybib-small.bib}

\end{document}